\documentclass[letterpaper, 10 pt, conference]{ieeeconf}  

\IEEEoverridecommandlockouts                              

\usepackage{multirow}
\usepackage{wrapfig}
\usepackage{subfigure}
\usepackage[utf8]{inputenc} 
\usepackage[T1]{fontenc}    

\usepackage{url}            
\usepackage{booktabs}       
\usepackage{amsfonts}       
\usepackage{nicefrac}       
\usepackage{microtype}      
\usepackage{xcolor}         
\usepackage[pdftex]{graphicx}

\usepackage{amsmath}
\usepackage{amssymb}

\usepackage{tabularx}
\usepackage{array}
\usepackage{booktabs}
\newcolumntype{Y}{>{\centering\arraybackslash}X}
\usepackage{makecell}

\usepackage[pagebackref,breaklinks,colorlinks]{hyperref}      

\usepackage[capitalize]{cleveref}
\crefname{section}{Sec.}{Secs.}
\Crefname{section}{Section}{Sections}
\Crefname{table}{Table}{Tables}
\crefname{table}{Tab.}{Tabs.}

\usepackage[table,xcdraw]{xcolor}
\definecolor{iccvblue}{rgb}{0.21,0.49,0.74}
\usepackage{algorithm}
\usepackage{algorithmic}

\usepackage[normalem]{ulem}
\useunder{\uline}{\ul}{}

\title{\LARGE \bf
MetaDAT: Generalizable Trajectory Prediction via Meta Pre-training and Data-Adaptive Test-Time Updating
}

\author{Yuning Wang$^{1}$, Pu Zhang$^{2}$, Yuan He$^{2}$, Ke Wang$^{2}$, Jianru Xue$^{1}$ 
\thanks{$^{1}$Yuning Wang and Jianru Xue are with the State Key Laboratory of Human-Machine Hybrid Augmented Intelligence, IAIR, Xi'an Jiaotong University, China. Corresponding author: Jianru Xue. \tt\small{wangyn@stu.xjtu.edu.cn. jrxue@mails.xjtu.edu.cn.} }
\thanks{$^{2}$Pu Zhang, Yuan He, and Ke Wang are with KargoBot, China. }
}

\begin{document}

\maketitle

\maketitle
\thispagestyle{empty}
\pagestyle{empty}

\begin{abstract}
Existing trajectory prediction methods exhibit significant performance degradation under distribution shifts during test time. Although test-time training techniques have been explored to enable adaptation, current approaches rely on an offline pre-trained predictor that lacks online learning flexibility. Moreover, they depend on fixed online model updating rules that do not accommodate the specific characteristics of test data. To address these limitations, we first propose a meta-learning framework to directly optimize the predictor for fast and accurate online adaptation, which performs bi-level optimization on the performance of simulated test-time adaptation tasks during pre-training. Furthermore, at test time, we introduce a data-adaptive model updating mechanism that dynamically adjusts the predefined learning rates and updating frequencies based on online partial derivatives and hard sample selection. This mechanism enables the online learning rate to suit the test data,  and focuses on informative hard samples to enhance efficiency. Experiments are conducted on various challenging cross-dataset distribution shift scenarios, including nuScenes, Lyft, and Waymo. Results demonstrate that our method achieves superior adaptation accuracy, surpassing state-of-the-art test-time training methods for trajectory prediction. Additionally, our method excels under suboptimal learning rates and high FPS demands, showcasing its robustness and practicality.
\end{abstract}

\section{Introduction}
Trajectory prediction plays a crucial role in understanding environments and building world models, making it a cornerstone of autonomous driving  \cite{houenou2013vehicle,hu2023planning}. Mainstream research \cite{zhou2022hivt,shi2022motion,cheng2023forecast,zhou2023query} focuses on building strong data-driven predictors on pre-collected datasets, which is referred to as \textit{offline training}. However, these predictors often struggle with distribution shifts in test data, such as changes in road structures \cite{ye2023improving}, interaction patterns \cite{xu2022adaptive}, and driving styles \cite{vasudevan2024planning}, leading to significant performance degradation. Such limitations pose critical safety risks and highlight the need for more robust solutions.

\begin{figure}[t]
\centering
\subfigure[]{
\centering
\includegraphics[width=0.85\linewidth]{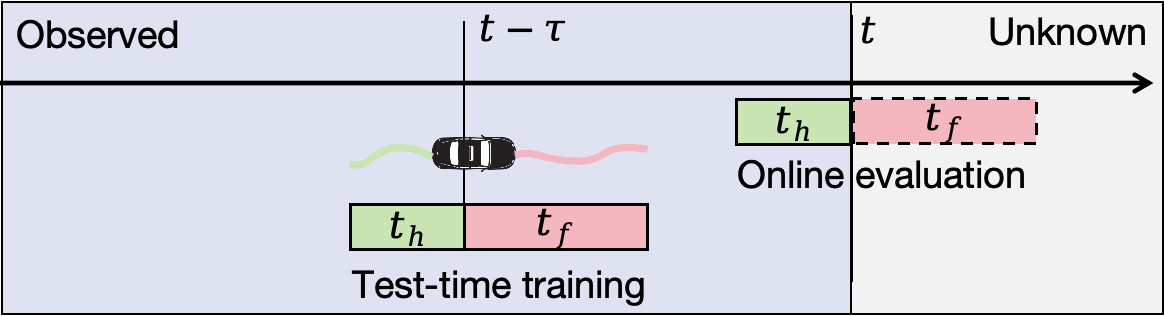}
}
\subfigure[]{
\centering
\includegraphics[width=0.95\linewidth]{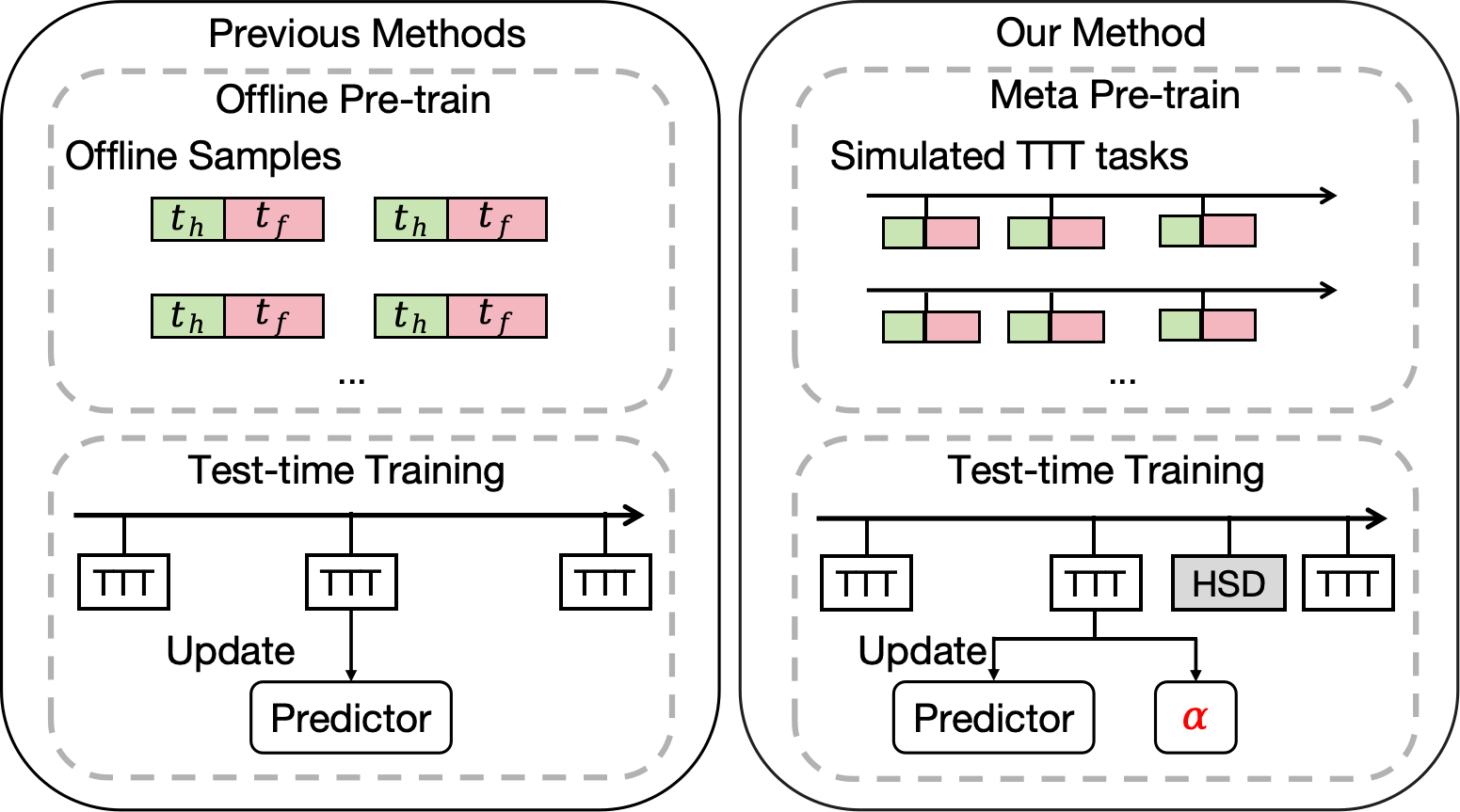}
}
\vspace{-0.2cm}
\caption{(a) Illustration for the test-time training and online evaluation protocol for trajectory prediction. 
(b) Previous offline pre-training overlooks the offline-online misalignment between pre-training and adaptation. In contrast, our method aligns these two phases via meta pre-training (MP). Additionally, instead of using fixed updating rules, we propose data-adaptive test-time updating by dynamic optimization of the learning rate $\alpha$ (DLO) and additional hard-sample-driven updates (HSD).
}
\vspace{-0.4cm}
\label{fig:ttt}
\end{figure}

In this paper, we investigate \textit{test-time training (TTT)} for trajectory prediction \cite{park2024t4p,ivanovic2023expanding,wang2021online} to handle distribution shifts during testing. Our approach adapts a model pre-trained on \textit{a source dataset} to an \textit{unseen target domain} by online learning on the test-time data. Thanks to the auto-labeling nature of the trajectory prediction task, the observations at the current time work as labels for previous predictions. The training and evaluation protocol of TTT is illustrated in \cref{fig:ttt}.

Existing trajectory predictors demonstrate constrained online learning effectiveness, primarily due to two critical issues. \textbf{Firstly, existing offline pre-training objectives misalign with online adaptation}. Current pre-training objectives focus on offline prediction accuracy for in-distribution samples but overlook online adaptation capability. As a result, the predictor adapts slowly, and the pre-trained representations quickly deteriorate. \textbf{Secondly, current fixed online updating strategies cannot suit the varying test-time data.}
An effective model updating rule should suit the test data characteristics. For example, the learning rate should be related to the magnitude of the distribution shift between the training and the unknown test data and be adaptively adjusted during learning. Also, efficient model updating should focus on the most representative samples for the distribution shift. Conventional methods, however, rely on rigid, predetermined learning rates and update frequencies and are unaware of the test data, severely limiting their performance and efficiency.

To tackle the aforementioned problem, we design a new TTT framework for trajectory prediction called MetaDAT: Meta pre-training and Data-Adaptive Test-time updating. Our framework introduces a bi-level optimization process in the pre-training stage based on meta-learning, to directly optimize a flexible starting point for test-time updating. We align the offline training with online objectives by simulating test-time training tasks on the source dataset, and optimizing a meta-objective across those tasks that evaluates the online adapting performance. For the test-time updating process, we propose a unique learning rate optimization based on the online partial derivatives. Additionally, a hard-sample-driven method is proposed to focus the model updates on the most critical samples. Our contributions can be summarized as follows.
 
\begin{itemize}

\item We introduce a meta pre-training framework for test-time training that addresses the offline-online misalignment during pre-training, fully leveraging the potential of the pre-training stage and initializing a flexible predictor for subsequent adaptation.

\item We propose a data-adaptive updating method for the test-time learning process, incorporating dynamic learning rates optimization and hard-sample-driven model updates, enabling effective and efficient adaptation.

\item Experiments across widely used datasets and different configurations demonstrate that our method achieves state-of-the-art performance in both accuracy and efficiency. Our method can also make improvements under suboptimal learning rates and few-shot cases.

\end{itemize}

\section{Related Work}

\subsection{Trajectory Prediction} 
Recently, data-driven trajectory predictors have achieved remarkable performance \cite{cheng2023forecast,zhou2023query,zhou2024smartrefine,salzmann2020trajectron++} by utilizing large-scale pre-collected datasets \cite{sun2020scalability,caesar2020nuscenes,zhan2019interaction,houston2021one} and advanced techniques such as transformer architectures \cite{cheng2023forecast,zhou2023query,park2023leveraging} and masked autoencoder (MAE) loss \cite{cheng2023forecast,chen2023traj}. Despite their high accuracy, research shows that the offline trained models are prone to test-time distribution shifts \cite{xu2022adaptive,ivanovic2023expanding,park2024t4p,vasudevan2024planning}. 
Some work investigates transfer learning between distributions by anticipating the test domains \cite{xu2022adaptive,ye2023improving}. Those methods are unreliable when the domain shifts differently from anticipation. In contrast, our work focuses on test-time training for trajectory prediction, adapting to unknown domain shifts without prior assumptions.

\subsection{Test-Time Training}
Test-time training updates the model in an online manner
on the newly observed data during testing. Recent TTT methods for image classification focus primarily on developing self-supervised losses to address the lack of labels during testing \cite{liu2021ttt++,gandelsman2022test,sun2020test,chen2022contrastive,mirza2023mate}. 
Trajectory prediction, however, has a unique auto-labeling nature, making TTT for this task a distinct problem with label supervision.  Several TTT methods specifically designed for prediction have been developed.  For instance, MEK \cite{wang2021online} employs an extended Kalman filter as the online optimizer, while T4P \cite{park2024t4p} incorporates an MAE loss and actor-specific tokens. None of the previous methods considered the offline-online misalignment. The most relevant method is AML \cite{ivanovic2023expanding}, which also utilizes meta-learning. However, AML only adapts the last Bayesian linear regression layer in the decoder, which limits the adaptation ability of deeper model representations. In contrast, we introduce a more universal meta pre-training framework to unleash the full model potential. Furthermore, we introduce a unique data-adaptive test-time updating technique. 

\begin{figure*}[h]
\centering
\includegraphics[width=0.85\linewidth]{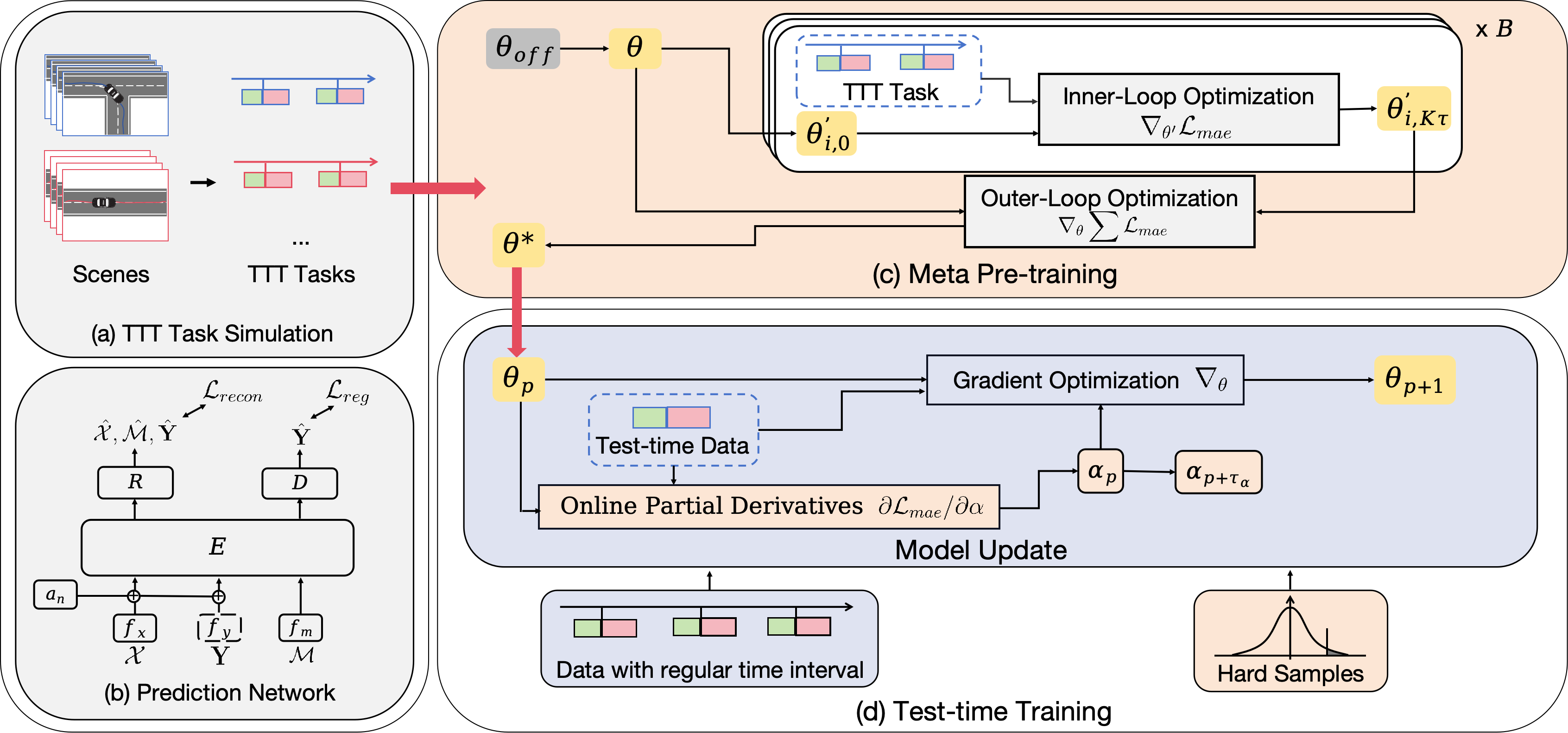}
\caption{Overview of our MetaDAT framework. MetaDAT first performs meta pre-training across the simulated test-time training tasks on the source dataset, then the pre-trained model goes through data-adaptive test-time training on the target dataset under distribution shifts. Newly proposed modules in our framework are highlighted in orange, while the TTT baseline is in purple.
}
\label{fig:overview}
\end{figure*}

\subsection{Meta-Learning For Trajectory Prediction}
Meta-learning focuses on making models learn how to learn for different tasks \cite{finn2017model, sun2019meta,chen2021meta,li2017meta,javed2019meta}. These methods have made significant progress in areas such as few-shot learning \cite{finn2017model,sun2019meta,chen2021meta} and domain adaptation \cite{li2020online,wang2021online}, which is crucial for the generalizability of autonomous driving. Recently, some studies have explored meta learning for trajectory prediction.  MENTOR \cite{pourkeshavarz2023learn} uses meta learning to perform multi-task learning, balancing auxiliary losses with the main prediction loss. MetaTraj \cite{shi2023metatraj}
uses meta-learning to perform zero-shot transfer learning across scenes. None of those methods considered solving the offline-online objective misalignment for test-time training of the predictor.

\section{Methodology}

\subsection{Problem Formulation}
\textbf{Trajectory Prediction.} Trajectory prediction is a sequential prediction problem. At time step $t$, given the observation $\mathbf{X}_t= \{\mathcal{X}_t, \mathcal{M}_t\}$, where $\mathcal{X}_t =\mathbf{x}^{0:N}_{t-t_h:t}$ is the $N$ agent trajectories over the past horizon $t_h$ and $\mathcal{M}_t$ is the surrounding map information, the goal is to narrow the gap between the predictions $\mathbf{\hat{Y}}_t$ and the ground truth locations of agents $\mathbf{Y}_t=\mathbf{y}^{0:N}_{t:t+t_f}$ in future horizon $t_f$. For simplicity, we omit $N$ in the following sections. In this work, we consider the situation where the test-time \textit{target data} $D^T=\{\mathbf{X}^{T},\mathbf{Y}^{T}\}$ does not share the same distribution as the training \textit{source data} $D^S=\{\mathbf{X}^{S},\mathbf{Y}^{S}\}$.

\textbf{Test-time Training And Online Evaluation.} We perform test-time training to adapt to $D^T$, and conduct online evaluations to assess adaptation performance, as illustrated in \cref{fig:ttt}.
We adopt the widely used online trajectory prediction setting \cite{park2024t4p,ivanovic2023expanding,wang2021online}, where the test data consists of multiple
\textit{scenes}, each containing \textit{temporally ordered data samples}, and one sample is observed at each time step as time passes. Given a time interval $\tau$, at time $t$ we can get the ground truth future $\mathbf{Y}_{t-\tau} =\mathbf{x}_{t-{\tau}:t-{\tau}+t_f}$ for previous prediction at time $t-\tau$, thus we can update the predictor with supervision data $\{\mathbf{X}_{t-\tau},\mathbf{Y}_{t-\tau}\}$. The online evaluation at time $t$ is performed on the prediction $\mathbf{Y}_t$ for current $\mathbf{X}_t$.

\subsection{Overview}
We propose a test-time training framework for trajectory prediction called MetaDAT. The overview of MetaDAT is illustrated in \cref{fig:overview}. MetaDAT consists of a \textit{meta pre-training} phase (\cref{fig:overview} (c)) and a \textit{data-adaptive test-time updating} phase (\cref{fig:overview} (d)). During model pre-training, pseudo TTT tasks are simulated through temporal scene splitting.
Meta-learning is employed on these tasks to derive a flexible model initialization 
$\theta^*$ optimized for subsequent updating. During test-time updating, we propose online dynamic optimization of the learning rate $\alpha$ and perform hard-sample-driven model updates, ensuring that the online learning process suits the test-time data and concentrates on the critical samples.

\subsection{Predictor Model}
For our prediction network in \cref{fig:overview}.(b), we adopt the SOTA predictor ForecastMAE \cite{cheng2023forecast} as the backbone, the same as previous work T4P \cite{park2024t4p} to enable fair comparison.  
ForecastMAE consists of embedding layers $\{f_x,f_y,f_m\}$, an encoder $E$, a decoder $D$, and an MAE reconstruction branch $R$. Note that the future input $\mathbf{Y}$ is only for training.

\textbf{Pre-training Stage.} We adopt the modification in T4P \cite{park2024t4p}: instead of using a two-stage training \cite{cheng2023forecast}, the model parameter $\theta=\{f,E,D,R\}$ is optimized jointly on the MAE loss $\mathcal{L}_{mae}$, which consists of the regression loss $\mathcal{L}_{reg}$ and the reconstruction loss $\mathcal{L}_{recon}$ :
\begin{equation}
\mathop{\min}_{\theta} \mathbb{E}_{D^S}  
\mathcal{L}_{mae}=\mathcal{L}_{reg}(\mathbf{X},\mathbf{Y})+\mathcal{L}_{recon}(\mathbf{X},\mathbf{Y})
\end{equation}

\textbf{Test-time Training.} At time step $t$, we update the model by minimizing the MAE loss at the previous time step $t-\tau$ given the TTT time interval $\tau$.
\begin{equation}
\begin{aligned}
\mathop{\min}_{\theta} \  
\mathcal{L}^{t-\tau}_{mae} &=\mathcal{L}_{reg}(\mathbf{X}_{t-\tau},\mathbf{Y}_{t-\tau})+\mathcal{L}_{recon}(\mathbf{X}_{t-\tau},\mathbf{Y}_{t-\tau})
\end{aligned}
\label{eq: online mae}
\end{equation}

We also adopt the actor-specific tokens proposed by T4P \cite{park2024t4p} to learn actor-wise habits. 
In the TTT stage, the actor-specific tokens $a_n$ are added to the embedding for each actor $n$ to obtain the actor-wise feature $h_x,h_y$, then $h_x,h_y$ are fed into the encoder $E$. 
\begin{equation}
h_x,h_y = f_x(\mathbf{X}) + a_n,f_y(\mathbf{Y}) + a_n
\end{equation}

\subsection{Meta Pre-training.}
We propose a meta pre-training (MP) framework to directly optimize the pre-trained model $\theta$ for test-time training on $D^{s}$, solving the offline-online misalignment.

\textbf{TTT Task Simulation.} To optimize the model for test-time adaptation, we first partition $D^{s}$ into separate sub-domains and simulate online test-time training tasks. For trajectory prediction, prior research has shown that different driving scenes naturally represent consistent sub-domains \cite{xu2022adaptive,park2024t4p, shi2023metatraj}, as each scene comprises unique agent behavioral patterns and distinct road structures. Thus, we regard driving scenes as sub-domains and simulate TTT tasks on them. Specifically, we divide the source dataset into individual driving scenes, and organize the samples within each scene $s$ of length $t_s$ in temporal order to create the online trajectory sequence $\mathbf{S}=\{\mathbf{X}_0,\mathbf{X}_1,...,\mathbf{X}_{t_s}\}$. This process generates an online sequence set $\mathcal{S}=\{\mathbf{S}_0,\mathbf{S}_1,...\}$ from all the scenes in $D^{s}$, and the TTT task set $\mathcal{T}=\{\mathbf{T}_0,\mathbf{T}_1,...\}$ is defined by online learning on $\mathcal{S}$.

\begin{algorithm}[t]
    \caption{Meta Pre-training}
    \begin{algorithmic}[1]
        \REQUIRE inner learning rate $\alpha_{in}$,
        meta-learning rate $\beta$, initial model weights $\theta \leftarrow \theta_{off}$, TTT task set $\mathcal{T}$, meta-learning batch size $B$, inner-loop step $K$, time interval $\tau$.
        \
        \WHILE{not converge} 
                \STATE sample TTT task batch $\{\mathbf{T}_0,...\mathbf{T}_B\} \sim \mathcal{T}$
                \FOR{each $\mathbf{S}_i$} 
                    \STATE initialize model parameter $\theta'_{i,0}=\theta$ 
                     \FOR{$t=\tau,2\tau...,K\tau$} 
                        \STATE Compute adapted parameters with gradient descent: $\theta'_{i,\tau}=\theta'_{i,t-\tau}-\alpha_{in}\nabla_{\theta'_{i,t-\tau}}\mathcal{L}^{i,t-\tau}_{mae}(\theta'_{i,t-\tau})$
                     \ENDFOR
                \ENDFOR
                \STATE Update meta-parameters:
                $\theta \leftarrow \theta-\beta\nabla_{\theta}\sum_i\mathcal{L}^{i,K\tau}_{mae}$
        \ENDWHILE
    \RETURN optimized model weights $\theta^*$
    \end{algorithmic}
    \label{alg:maml}
\end{algorithm}

\textbf{Meta-Learning.} Given the initial model parameter $\theta$ and the TTT task set $\mathcal{T}$, we perform a bi-level optimization \cite{finn2017model}: in the inner loop, the model $\theta$ undergoes simulated online test-time training to yield $\theta'$; in the outer loop, a meta-objective evaluates the adaptation performance and optimizes the initial parameters $\theta$. This process is illustrated in \cref{alg:maml}. We first sample a task batch $\{\mathbf{T}_0,...\mathbf{T}_B\}$ with batch size $B$. For a single online TTT task $\mathbf{T}_i$, we initialize task-specific model parameters $\theta'_{i,0}=\theta$ at $t=0$, then $\theta'_i$ is optimized by performing $K$ step test-time training in the inner loop. For each training step $t=\tau,2\tau...,K\tau$,
\begin{equation}
\theta'_{i,\tau}=\theta'_{i,t-\tau}-\alpha_{in}\nabla_{\theta'_{i,t-\tau}}\mathcal{L}^{i,t-\tau}_{mae}(\theta'_{i,t-\tau}).
\end{equation}
The adaptation performance is evaluated by the prediction loss $\mathcal{L}^{i,K\tau}_{mae}$ after the last adaptation step.
After $B$ inner loops, we perform meta optimization across the $B$ evaluation loss $\mathcal{L}^{i,K\tau}_{mae}$s with respect to the initial parameter $\theta$ in the outer loop. The optimization on $\theta$ is as follow:
\begin{equation}
    \theta = \theta-\beta\nabla_{\theta}\sum_i\mathcal{L}^{i,K\tau}_{mae}
\end{equation}
This equation contains second-order gradients with Hessian-vector products, which is computation consuming. In practice, we use the first-order-approximation \cite{nichol2018first}s.

\textbf{Model Initialization.} The inner step of meta-learning makes the training process very time-consuming. Also, prior work \cite{chen2021meta} reveals that performing meta-learning over an offline pre-trained model can result in a model with better generalization. Therefore, instead of training from scratch, we initialize the model with offline pre-trained parameters $\theta_{off}$, $\theta=\theta_{off}$ before meta-learning.

\subsection{Data Adaptive Test-time Updating.}

In addition to a good initialization, the test-time model updating rule is also critical to the adaptation performance. Conventional online updating methods adopt fixed learning rates and update frequencies, which fails to account for the unknown characteristics of test data.
Recognizing this issue,  we propose a data-adaptive mechanism for test-time updating.

\subsubsection{Dynamic Learning Rate Optimization (DLO)} We employ online partial derivatives to dynamically optimize the learning rate according to the observed performance. For simplicity, we consider stochastic gradient descent for parameter $\theta_p$ at the $p$th TTT step with learning rate $\alpha$, 
\begin{equation}
 \theta_{p}=\theta_{p-1}-\alpha\nabla_{\theta}\mathcal{L}_{mae}(\theta_{p-1}),
 \label{eq:sgd}
\end{equation}

It is straightforward to extend to more complex optimizers such as AdamW \cite{loshchilov2017decoupled}.

 Inspired by \cite{baydin2017online}, taking the assumption that the optimal $\alpha$ does not change much between two consecutive updates, we can use the partial derivative $\partial \mathcal{L}_{mae}(\theta_{p-1})/\partial \alpha$ on step $p-1$ to optimize $\alpha$ on step $p$. Noting $ \theta_{p-1}=\theta_{p-2}-\alpha\nabla_{\theta}\mathcal{L}_{mae}(\theta_{p-2})$ and applying the chain rule, we can get:
\begin{equation}
\begin{aligned}
\frac{\partial \mathcal{L}_{mae}(\theta_{p-1})}{\partial \alpha} &=\nabla_{\theta}\mathcal{L}_{mae}(\theta_{p-1})  \cdot\\
& \frac{\partial (\theta_{p-2}-\alpha\nabla_{\theta}\mathcal{L}_{mae}(\theta_{p-2}))}{\partial \alpha} \\
&=\nabla_{\theta}\mathcal{L}_{mae}(\theta_{p-1})\cdot(-\nabla_{\theta}\mathcal{L}_{mae}(\theta_{p-2}))
\end{aligned}
\end{equation}
And then $\alpha$ is updated as follows:
\begin{equation}
\begin{aligned}
\alpha_{p} &=\alpha_{p-1}-\gamma\partial \mathcal{L}_{mae}(\theta_{p-1})/\partial \alpha \\
&= \alpha_{p-1}+\gamma\nabla_{\theta}\mathcal{L}_{mae}(\theta_{p-1})\cdot\nabla_{\theta}\mathcal{L}_{mae}(\theta_{p-2}),
\end{aligned}
\end{equation}
where $\gamma$ is the learning rate for $\alpha$.

In practice, we update the learning rate by averaging over a longer interval $\tau_{\alpha}$ to stabilize the training process:
\begin{equation}
\alpha_{p} = \alpha_{p-\tau_{\alpha}}+\gamma\nabla_{\theta}\mathcal{L}_{mae}(\theta_{p-1})\cdot \frac{1}{\tau_{\alpha}}\sum^{p-2}_{j=p-\tau_{\alpha}-1}\nabla_{\theta}\mathcal{L}_{mae}(\theta_{j}),
\end{equation}

We adopt different $\alpha$s for each network layer, with the same initial learning rate $\alpha_{init}$. This can better liberate the network flexibility of the adaptive process without the additional need for manual hyperparameter tuning.

\subsubsection{Hard-sample-driven Model Updates (HSD)} Autonomous driving data exhibits a long-tail distribution, where a small fraction of challenging data—such as scenarios involving intensive interactions or heavy reliance on road maps—is particularly critical and demands sophisticated modeling \cite{makansi2021exposing,wang2023fend}. These data are most susceptible to performance degradation under distribution shift and best represent the information that needs to be learned in the current domain. We identify those critical hard samples by comparing the prediction error $e$ with a running error mean $m$ and a running error standard deviation $\sigma$ as follows, and perform additional updates on them:
\begin{equation}
e> m+k\sigma.
\label{eq:error update}
\end{equation}
 Equation \ref{eq:error update} restricts the updates to only a small number of samples. Therefore, we can utilize hard-sample-driven updates without sacrificing efficiency.  Learning rate optimization is not implemented on these hard samples.

\section{Experiments}

\begin{table*}[]
\centering
\caption{Performance on various distribution shifts. The predictor is trained on the source dataset, and test-time trained and online evaluated on the target dataset (Source \textrightarrow Target). * means reproduced results by ourselves. }
\footnotesize
\begin{tabular}{l|lll|l|ll|l}
\hline
                                                                       & \multicolumn{3}{l|}{Short-term prediction (1/3/0.1)}                                                          &                      & \multicolumn{2}{l|}{long-term prediction (2/6/0.5)}                       &                   \\ \cline{2-8} 
\multirow{-2}{*}{\begin{tabular}[c]{@{}l@{}} $mADE_6$\\ /$mFDE_6$\end{tabular}} & Lyft\textrightarrow nuS & nuS\textrightarrow Way & Way\textrightarrow nuS & Mean                 & nuS\textrightarrow Lyft & Lyft\textrightarrow nuS & Mean              \\ \hline
Joint Training                                                         & 0.506/1.102                         & 0.472/1.125                        & 0.374/0.718                        & 0.450/0.982          & 1.108/2.597                         & 1.398/3.109                         & 1.253/2.853       \\
DUA                                                                    & 0.474/1.118                         & 0.516/1.294                        & 0.396/0.781                        & 0.462/1.064          & 1.365/3.257                         & 1.585/3.607                         & 1.475/3.432       \\
TENT(w/ sup)                                                           & 0.462/1.009                         & 0.448/1.071                        & 0.358/0.695                        & 0.423/0.925          & 1.068/2.514                         & 1.396/3.102                         & 1.232/2.808       \\
MEK                                                                    & 0.508/1.806                         & 0.405/1.061                        & 0.373/0.713                        & 0.429/1.192          & 1.006/2.369                         & 1.426/3.119                         & 1.216/2.744       \\
AML                                                                    & 0.454/1.954                         & 0.764/1.791                        & 0.483/0.962                        & 0.567/1.569          & 1.462/2.573                         & 1.698/3.367                         & 1.580/2.970       \\
T4P                                                                    & {\ul 0.357/0.770}                   & {\ul 0.336}/0.807                  & 0.323/0.656                        & {\ul0.339}/0.744          & 0.776/1.820                         & 1.254/2.802                         & 1.014/2.311       \\
T4P*                                                                   & 0.408/0.847                         &  0.343/{\ul 0.792}                  & {\ul 0.284/0.585}                  & 0.345/{\ul 0.741}    & {\ul 0.711/1.578}                   & {\ul 1.260/2.742}                   & {\ul 0.985/2.160} \\
\rowcolor[HTML]{EFEFEF} 
\textbf{MetaDAT(Ours)}                                                 & \textbf{0.332/0.683}                & \textbf{0.305/0.712}               & \textbf{0.266/0.548} & \textbf{0.301/0.648} & \textbf{0.648/1.472}                         & \textbf{1.177/2.551}                         & \textbf{0.912/2.011}       \\ \hline
\end{tabular}
\label{tab:quanti}
\end{table*}

\begin{table*}[H]
\centering
\caption{Performance on various distribution shifts. The predictor is trained on the source dataset, and test-time trained and online evaluated on the target dataset. * means reproduced results by ourselves.}
\label{tab:metadat_quanti}
\footnotesize
\begin{tabularx}{\linewidth}{@{}X *{3}{Y} Y *{2}{Y} Y@{}}
\toprule
\makecell[t]{$\rm mADE_6$ \\ $\rm mFDE_6$} & \multicolumn{3}{c}{aaa} & \multirow{2}{*}{Mean} & \multicolumn{2}{c}{bbb} & \multirow{2}{*}{Mean} \\
\cmidrule(lr){2-4} \cmidrule(lr){6-7}
& L $\to$ N & N$\to$ W & W$\to$ N & & N$\to$ L & L$\to$ N & \\
\midrule
Pretraing & 0.506/1.102 & 0.472/1.125 & 0.374/0.718 & 0.450/0.982 & 1.108/2.597 & 1.398/3.109 & 1.253/2.853 \\
DUA & 0.474/1.118 & 0.516/1.294 & 0.396/0.781 & 0.462/1.064 & 1.365/3.257 & 1.585/3.607 & 1.475/3.432 \\
TENT  & 0.462/1.009 & 0.448/1.071 & 0.358/0.695 & 0.423/0.925 & 1.068/2.514 & 1.396/3.102 & 1.232/2.808 \\
MEK & 0.508/1.806 & 0.405/1.061 & 0.373/0.713 & 0.429/1.192 & 1.006/2.369 & 1.426/3.119 & 1.216/2.744 \\
AML & 0.454/1.954 & 0.764/1.791 & 0.483/0.962 & 0.567/1.569 & 1.462/2.573 & 1.698/3.367 & 1.580/2.970 \\
T4P & \underline{0.357/0.770} & \underline{0.336}/0.807 & 0.323/0.656 & \underline{0.339}/0.744 & 0.776/1.820 & 1.254/2.802 & 1.014/2.311 \\
T4P* & 0.408/0.847 & 0.343/\underline{0.792} & \underline{0.284/0.585} & 0.345/\underline{0.741} & \underline{0.711/1.578} & \underline{1.260/2.742} & \underline{0.985/2.160} \\
\textbf{MetaDAT} & \textbf{0.332/0.683} & \textbf{0.305/0.712} & \textbf{0.266/0.548} & \textbf{0.301/0.648} & \textbf{0.648/1.472} & \textbf{1.177/2.551} & \textbf{0.912/2.011} \\
\bottomrule
\end{tabularx}
\end{table*}

\begin{table}[]
\caption{Comparison on more multi-modal prediction performance.}
\small
\centering
\begin{tabular}{l|lll}
\hline
\multirow{2}{*}{\begin{tabular}[c]{@{}l@{}}$mADE_1$\\ /$MR_6$\end{tabular}} & \multicolumn{3}{l}{Short-term prediction}                            \\ \cline{2-4} 
                                                                     & Lyft\textrightarrow nuS            &   nuS\textrightarrow Way          & Way\textrightarrow nuS               \\ \hline
T4P                                                                  &   1.050/0.088            & 0.805/0.091   &  0.911/0.037        \\
MetaDAT                                                                & \textbf{0.856/0.060} & \textbf{0.708/0.078}& \textbf{0.785/0.026} \\ \hline
\end{tabular}
\label{tab:more quanti}
\end{table}

\subsection{Datasets}
We conduct experiments on well-known trajectory datasets: Waymo \cite{sun2020scalability},nuScenes \cite{caesar2020nuscenes}, Lyft \cite{houston2021one}, to construct various cross-dataset distribution shifts. We make sure that scene lengths in the source dataset are enough to perform 
a single-step test-time update for meta pre-training. Trajdata \cite{ivanovic2024trajdata} is used as the dataset interface. Two kinds of widely used configurations are adopted: long-term prediction and short-term prediction. For short-term prediction, we watch 1 second and predict 3 seconds with a time interval of 0.1. For long-term prediction, we watch 2 seconds and predict 6 seconds with a time interval of 0.5. We evaluate the best-of-6 metrics \cite{chang2019argoverse} for multi-modal prediction: $mADE_6$ and $mFDE_6$. We also evaluate $mADE_1$ and $MR_6$ for comprehensive analysis.

\subsection{Baselines.}
We compare with widely used unsupervised/supervised test-time-training methods: DUA \cite{mirza2022norm} and TENT \cite{wang2020tent} with supervision, as-well-as state-of-the-art TTT methods specifically for trajectory prediction: MEK \cite{wang2021online}, AML \cite{ivanovic2023expanding}, and T4P \cite{park2024t4p}. All methods use the same network architecture as ours. We also add a \textbf{non-TTT baseline}: Joint-training, which means only pre-training the model on the source dataset using the joint MAE loss  $\mathcal{L}_{mae}$ and performing no test-time update.

\subsection{Implement Details.}
We use the same model structure as ForecastMAE \cite{cheng2023forecast,park2024t4p}. For offline pre-training, we adopt the same training schedule as T4P \cite{park2024t4p}. The offline pre-trained model works as model initialization for meta pre-training. For meta pre-training, we train for 8 epochs with a meta batch size $B=4$ using the AdamW \cite{loshchilov2017decoupled} optimizer with weight decay of 0.001. The inner-loop step $K$ is set as 4.  The learning rate $\beta$ for meta pre-training is set as 5e-4 initially and is gradually decayed by a cosine scheduler to 1e-6. The inner learning rate $\alpha_{in}$ is the same as the online learning rate $\alpha_{init}$. For the test-time training process, we select the initial learning rate  $\alpha_{init}$ as 0.001 for nuS \textrightarrow Way short-term and all the long-term experiments, and 0.01 for all other experiments. The AdamW optimizer \cite{loshchilov2017decoupled} is adopted for test-time training. We select the TTT time interval $\tau=t_f$ to allow the past GT future to contain the full prediction horizon. $\gamma$ is set as 1e-4, and the learning rate update interval $\tau_{\alpha}$ is 8. $k$ is set as 3. In \cref{sec:discuss}, the FPS results are calculated on a single NVIDIA-L4 GPU. 
 
\subsection{Results And Comparisons.}
 \textbf{Quantitative Results.} Table \ref{tab:quanti} summarizes our experimental results under various cross-dataset distribution shifts. All the results are averaged over 3 independent runs. Our method stably outperforms all comparison methods across various distribution shifts and different prediction configurations, showing the superiority and generalization ability of MetaDAT. Specifically, we outperform the second-best TTT method T4P \cite{park2024t4p} by 12.7\% on $mADE_6$ and 12.5\% on $mFDE_6$ under short-term prediction configuration, thanks to our meta pre-training to get a flexible model initialization, and our effective and efficient data-adaptive updating. Moreover, our method outperforms another meta-learning-based competitor AML \cite{ivanovic2023expanding} significantly. This highlights that our method has the flexibility to pre-train and finetune deeper model representations. We also compare with the second-best T4P on more metrics in \cref{tab:more quanti}, to better show our superior multi-modal prediction performance.

\begin{figure*}
\centering
\begin{subfigure}
\centering
\includegraphics[width=0.55\linewidth]{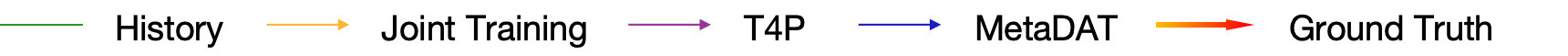} 
\end{subfigure}
\vspace{-0.4cm}
\begin{subfigure}
\centering
\includegraphics[width=0.75\linewidth]{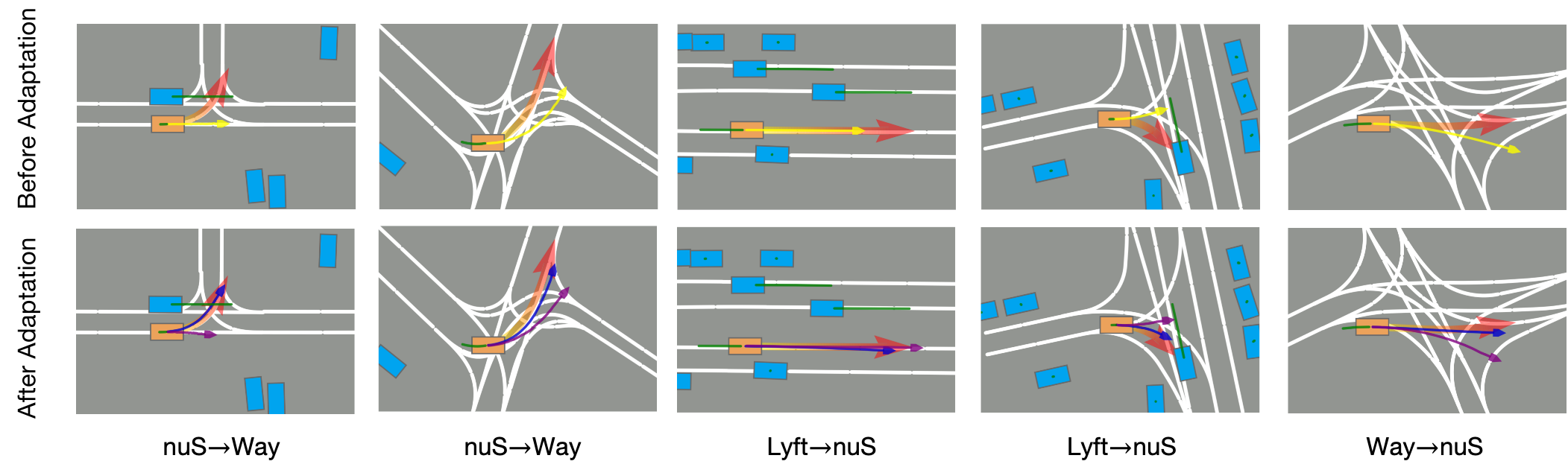}
\end{subfigure}
\caption{Qualitative visualizations on test-time training results. The top row shows the prediction results without adaptation, while the bottom row indicates the adaptation results using test-time training. We only visualize the best modality for multi-modal predictions.}
\label{fig:quali}
\end{figure*}

\begin{figure}
\centering
\includegraphics[width=0.65\linewidth]{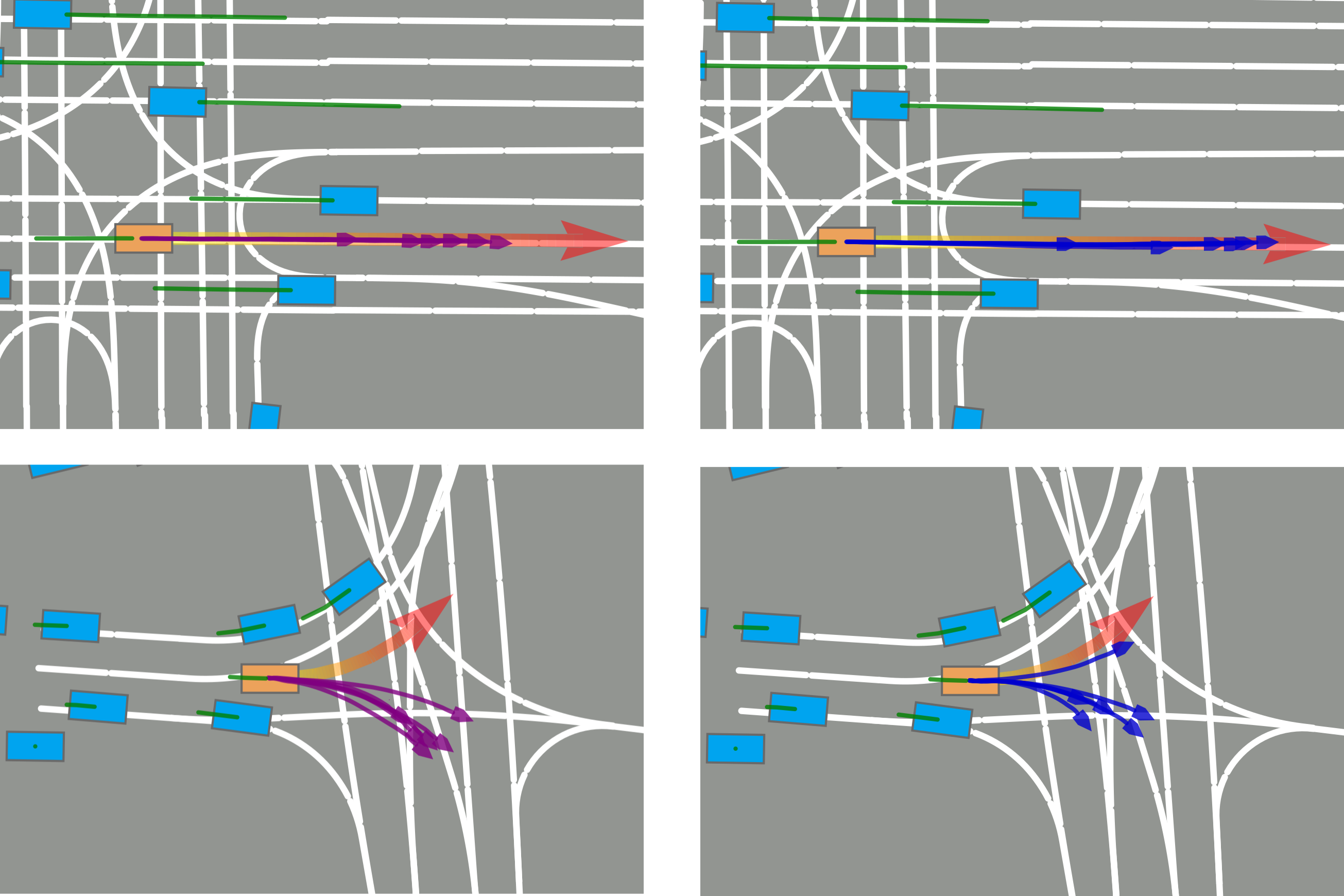}
\caption{Qualitative visualizations on multi-modal prediction results for TTT methods. Left: T4P. Right: MetaDAT.}
\label{fig:quali mm}
\end{figure}

 \textbf{Qualitative Results.} In Figure \ref{fig:quali} we visualize the qualitative prediction results after test-time training on various driving scenarios, such as going straight, turning, and crossing intersections.  We compare against two methods: Joint-training which performs no adaptation, and the previous SOTA TTT method T4P. Our method shows stable superior adaptation performance, surpassing both methods across various datasets. This demonstrates the effectiveness of the strategies we proposed in both pre-training and online training. Furthermore, qualitative comparison on the multi-modal prediction results in \cref{fig:quali mm} shows that our method can enhance both horizontal and vertical prediction modality diversities.

\subsection{Ablation And Discussions}
\label{sec:discuss}

\textbf{Ablations.}
In \cref{tab:ablation} we do ablations on different components of MetaDAT. The table illustrates that all three proposed modules: meta pre-training, dynamic learning rate optimization, and hard-sample-driven updates can lead to performance improvements when working individually. Combining all three modules leads to further improvements. This illustrates the complementarity of those modules, as they simultaneously focus on two different aspects: pre-training a powerful model initialization, and more adaptive test-time training by adjusting learning rate and updating frequency.

\begin{table}[]
\centering
\footnotesize
\caption{Ablations on different modules of MetaDAT.}
\begin{tabular}{lll|l|l|l}
\hline
\multicolumn{3}{l|}{Components}      & Short-term           & Long-term            & \multirow{2}{*}{Mean} \\ \cline{1-5}
MP         & DLO        & HSD        & Lyft\textrightarrow nuS             & nuS\textrightarrow Lyft             &                       \\ \hline
           &            &            & 0.408/0.847          & 0.711/1.578          & 0.560/1.213           \\
\checkmark &            &            & 0.355/0.734          & 0.672/1.491          & 0.514/1.112           \\
           & \checkmark &            & 0.376/0.776          & 0.684/1.538          & 0.530/1.157           \\
           &            & \checkmark & 0.400/0.836          & 0.707/1.552          & 0.554/1.194           \\
\checkmark & \checkmark &            & 0.347/0.702          & 0.650/\textbf{1.468}          & 0.498/1.085           \\
\checkmark & \checkmark & \checkmark & \textbf{0.332/0.683} & \textbf{0.648}/1.472 & \textbf{0.490/1.077}  \\ \hline
\end{tabular}
\label{tab:ablation}
\end{table}

\begin{table}[]
\centering
\caption{Robustness to learning rate $\alpha$($\alpha_{init}$) on nuS\textrightarrow Way short-term ($mADE_6/mFDE_6$).}
\begin{tabular}{l|lll}
\hline
        & $\alpha$=0.01 & $\alpha$=0.001 & $\alpha$=0.0001 \\ \hline
T4P     & 0.518/1.097 & 0.343/0.792  & 0.393/0.967   \\
T4P+DLO & 0.452/0.957 & 0.337/0.762  & 0.350/0.792   \\
MetaDAT & \textbf{0.407/0.887} & \textbf{0.305/0.712}  & \textbf{0.341/0.760}   \\ \hline
\end{tabular}
\label{tab: learning rate}
\end{table}

\textbf{Robustness To Different Learning Rate.} 
Robustness is a critical factor in online learning for trajectory prediction, as it significantly impacts the safety of real-time systems.
We studied the robustness of the test-time training algorithms to the initial online learning rates $\alpha$($\alpha_{init}$) in \cref{tab: learning rate}.  We can see that $\alpha$ largely influences the performance of T4P. Our proposed dynamic learning rate optimization can adaptively adjust $\alpha$ to suit the test data, therefore the performance under bad initial $\alpha$ is improved (T4P+DLO). Our full MetaDAT makes more progress, which might be because the meta pre-training can get a robust model initialization.

\begin{figure}
\centering
\subfigure{
\centering
\includegraphics[width=0.70\linewidth]{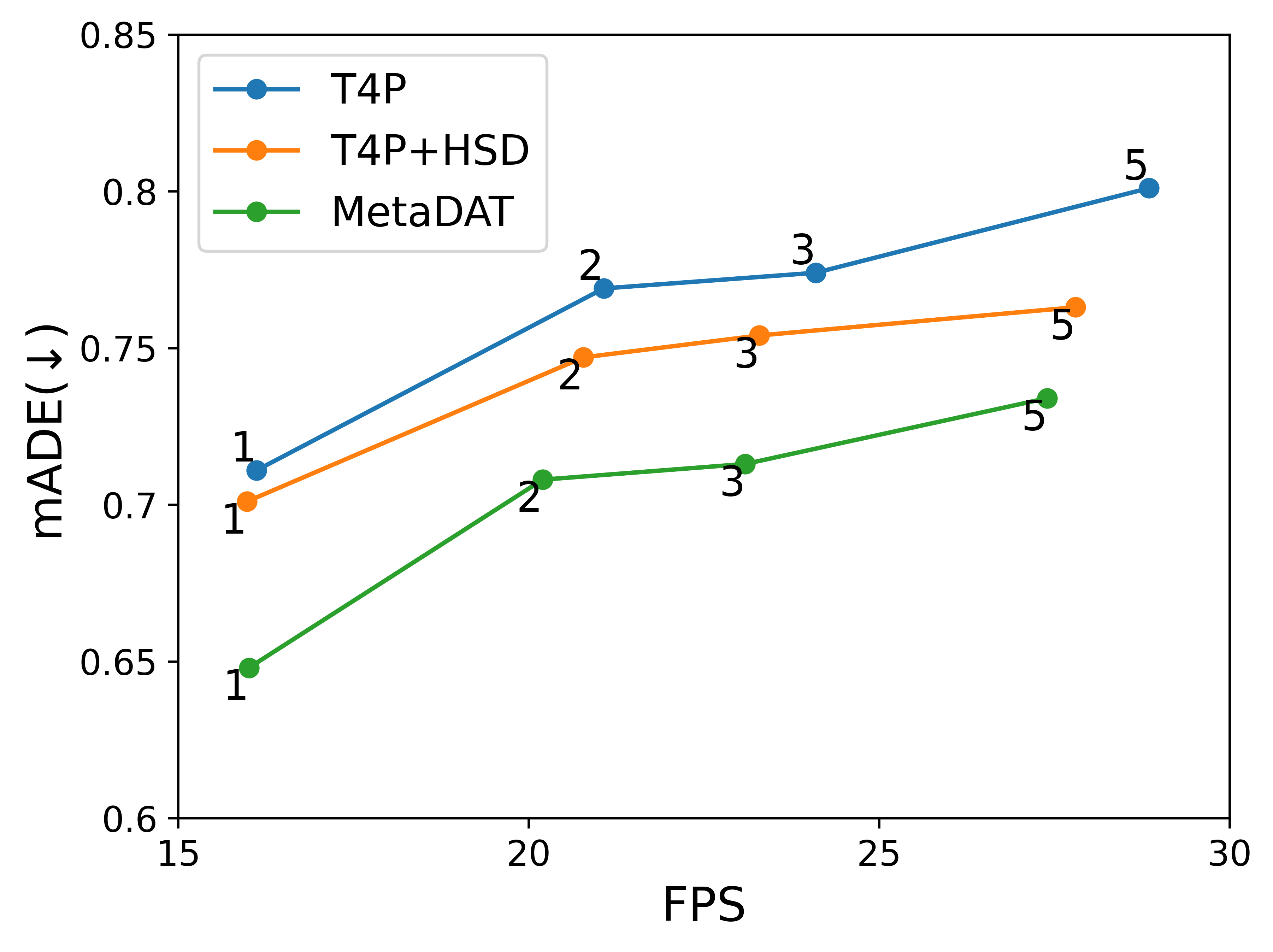} }
\vspace{-0.1cm}
\caption{Prediction accuracy and execution FPS on nuS\textrightarrow Lyft long-term experiments. The number next to each data point indicates the updating frequency. We plot results for frequencies of 1,2,3,5. MetaDAT demonstrates better prediction performance at the same FPS.}
\vspace{-0.3cm}
\label{fig:fps}
\end{figure}

\begin{table}[]
\centering
\caption{Few-shot adaptation performance ($mADE_6/mFDE_6$) on nuS\textrightarrow Way short-term.}
\begin{tabular}{l|lll}
\hline
Sample           & 2000                 & 5000                 & 10000                \\ \hline
T4P            & 0.350/0.849          & 0.348/0.801          & 0.343/0.792          \\
MetaDAT & \textbf{0.327/0.743} & \textbf{0.317/0.731} & \textbf{0.305/0.712} \\ \hline
\end{tabular}
\label{tab:few-shot}
\end{table}

\textbf{Efficiency.} Efficiency is another crucial factor for online learning.
 In \cref{fig:fps} we evaluate the frame per second (FPS) along with accuracy ($mADE_6$) of MetaDAT against the second-best method T4P.  Both methods can improve FPS by adjusting the update frequency: a frequency of 1 indicates updating at every opportunity, making $\tau=12$, while 2 means $\tau=24$. The HSD updating frequency is always 1. MetaDAT can improve the performances with little additional latency introduced, thanks to the hard-sample-driven updates.

\textbf{Few-shot Adaptation Results.} In \cref{tab:few-shot}
we analyze the prediction performance with different sample amounts for test-time adaptation. The table clearly demonstrates that MetaDAT consistently outperforms T4P across various data amounts. Notably, MetaDAT achieves impressive performance even with a small number of samples, such as 2000, highlighting its few-shot adaptation capability.

\textbf{Limitations And Future Work.} Our test-time training approach relies on accurate online detection and tracking to obtain accurate observed trajectories for training. However, in practice, noisy trajectories from imperfect perception can degrade model performance \cite{park2024improving,ivanovic2022propagating}. Thus, developing robust online learning strategies remains a key challenge.

\section{Conclusion}
We propose a Meta pre-training and Data Adaptive Test-time updating framework for trajectory predictors under distribution shift. Both quantitative and qualitative results demonstrate that our method achieves SOTA adaptation performance on various 
cross-dataset shifts and prediction configurations. Additionally, our approach enhances adaptation performance under few-shot conditions, suboptimal learning rates, and high-efficiency requirements, making it well-suited for real-world applications.

\section*{Acknowledgement}
This work is supported by the National Key Research and Development Program of China (Grant No. 2024YFE0210700)

{\small
\bibliographystyle{IEEEtran}
\bibliography{icra2}

@inproceedings{pourkeshavarz2023learn,
  title={Learn tarot with mentor: A meta-learned self-supervised approach for trajectory prediction},
  author={Pourkeshavarz, Mozhgan and Chen, Changhe and Rasouli, Amir},
  booktitle={Proceedings of the IEEE/CVF International Conference on Computer Vision},
  pages={8384--8393},
  year={2023}
}

@inproceedings{chen2022contrastive,
  title={Contrastive test-time adaptation},
  author={Chen, Dian and Wang, Dequan and Darrell, Trevor and Ebrahimi, Sayna},
  booktitle={Proceedings of the IEEE/CVF Conference on Computer Vision and Pattern Recognition},
  pages={295--305},
  year={2022}
}

@inproceedings{wang2023fend,
  title={Fend: A future enhanced distribution-aware contrastive learning framework for long-tail trajectory prediction},
  author={Wang, Yuning and Zhang, Pu and Bai, Lei and Xue, Jianru},
  booktitle={Proceedings of the IEEE/CVF conference on computer vision and pattern recognition},
  pages={1400--1409},
  year={2023}
}

@inproceedings{makansi2021exposing,
  title={On exposing the challenging long tail in future prediction of traffic actors},
  author={Makansi, Osama and {\c{C}}i{\c{c}}ek, {\"O}zg{\"u}n and Marrakchi, Yassine and Brox, Thomas},
  booktitle={Proceedings of the IEEE/CVF International Conference on Computer Vision},
  pages={13147--13157},
  year={2021}
}

@article{liu2021ttt++,
  title={Ttt++: When does self-supervised test-time training fail or thrive?},
  author={Liu, Yuejiang and Kothari, Parth and Van Delft, Bastien and Bellot-Gurlet, Baptiste and Mordan, Taylor and Alahi, Alexandre},
  journal={Advances in Neural Information Processing Systems},
  volume={34},
  pages={21808--21820},
  year={2021}
}

@inproceedings{sun2020test,
  title={Test-time training with self-supervision for generalization under distribution shifts},
  author={Sun, Yu and Wang, Xiaolong and Liu, Zhuang and Miller, John and Efros, Alexei and Hardt, Moritz},
  booktitle={International conference on machine learning},
  pages={9229--9248},
  year={2020},
  organization={PMLR}
}

@article{gandelsman2022test,
  title={Test-time training with masked autoencoders},
  author={Gandelsman, Yossi and Sun, Yu and Chen, Xinlei and Efros, Alexei},
  journal={Advances in Neural Information Processing Systems},
  volume={35},
  pages={29374--29385},
  year={2022}
}

@inproceedings{mirza2023mate,
  title={Mate: Masked autoencoders are online 3d test-time learners},
  author={Mirza, M Jehanzeb and Shin, Inkyu and Lin, Wei and Schriebl, Andreas and Sun, Kunyang and Choe, Jaesung and Kozinski, Mateusz and Possegger, Horst and Kweon, In So and Yoon, Kuk-Jin and others},
  booktitle={Proceedings of the IEEE/CVF International Conference on Computer Vision},
  pages={16709--16718},
  year={2023}
}

@article{shi2023metatraj,
  title={MetaTraj: meta-learning for cross-scene cross-object trajectory prediction},
  author={Shi, Xiaodan and Zhang, Haoran and Yuan, Wei and Shibasaki, Ryosuke},
  journal={IEEE Transactions on Intelligent Transportation Systems},
  year={2023},
  publisher={IEEE}
}

@article{shi2022motion,
  title={Motion transformer with global intention localization and local movement refinement},
  author={Shi, Shaoshuai and Jiang, Li and Dai, Dengxin and Schiele, Bernt},
  journal={Advances in Neural Information Processing Systems},
  volume={35},
  pages={6531--6543},
  year={2022}
}

@inproceedings{cheng2023forecast,
  title={Forecast-mae: Self-supervised pre-training for motion forecasting with masked autoencoders},
  author={Cheng, Jie and Mei, Xiaodong and Liu, Ming},
  booktitle={Proceedings of the IEEE/CVF International Conference on Computer Vision},
  pages={8679--8689},
  year={2023}
}

@article{wang2021online,
  title={Online adaptation of neural network models by modified extended kalman filter for customizable and transferable driving behavior prediction},
  author={Wang, Letian and Hu, Yeping and Liu, Changliu},
  journal={arXiv preprint arXiv:2112.06129},
  year={2021}
}

@inproceedings{li2020online,
  title={Online meta-learning for multi-source and semi-supervised domain adaptation},
  author={Li, Da and Hospedales, Timothy},
  booktitle={European Conference on Computer Vision},
  pages={382--403},
  year={2020},
  organization={Springer}
}

@inproceedings{sun2019meta,
  title={Meta-transfer learning for few-shot learning},
  author={Sun, Qianru and Liu, Yaoyao and Chua, Tat-Seng and Schiele, Bernt},
  booktitle={Proceedings of the IEEE/CVF conference on computer vision and pattern recognition},
  pages={403--412},
  year={2019}
}

@inproceedings{xu2022adaptive,
  title={Adaptive trajectory prediction via transferable gnn},
  author={Xu, Yi and Wang, Lichen and Wang, Yizhou and Fu, Yun},
  booktitle={Proceedings of the IEEE/CVF conference on computer vision and pattern recognition},
  pages={6520--6531},
  year={2022}
}

@inproceedings{chen2023traj,
  title={Traj-mae: Masked autoencoders for trajectory prediction},
  author={Chen, Hao and Wang, Jiaze and Shao, Kun and Liu, Furui and Hao, Jianye and Guan, Chenyong and Chen, Guangyong and Heng, Pheng-Ann},
  booktitle={Proceedings of the IEEE/CVF International Conference on Computer Vision},
  pages={8351--8362},
  year={2023}
}

@article{baydin2017online,
  title={Online learning rate adaptation with hypergradient descent},
  author={Baydin, Atilim Gunes and Cornish, Robert and Rubio, David Martinez and Schmidt, Mark and Wood, Frank},
  journal={arXiv preprint arXiv:1703.04782},
  year={2017}
}

@article{ivanovic2024trajdata,
  title={trajdata: A unified interface to multiple human trajectory datasets},
  author={Ivanovic, Boris and Song, Guanyu and Gilitschenski, Igor and Pavone, Marco},
  journal={Advances in Neural Information Processing Systems},
  volume={36},
  year={2024}
}

@article{loshchilov2017decoupled,
  title={Decoupled weight decay regularization},
  author={Loshchilov, I},
  journal={arXiv preprint arXiv:1711.05101},
  year={2017}
}

@inproceedings{chen2021meta,
  title={Meta-baseline: Exploring simple meta-learning for few-shot learning},
  author={Chen, Yinbo and Liu, Zhuang and Xu, Huijuan and Darrell, Trevor and Wang, Xiaolong},
  booktitle={Proceedings of the IEEE/CVF international conference on computer vision},
  pages={9062--9071},
  year={2021}
}

@inproceedings{mirza2022norm,
  title={The norm must go on: Dynamic unsupervised domain adaptation by normalization},
  author={Mirza, M Jehanzeb and Micorek, Jakub and Possegger, Horst and Bischof, Horst},
  booktitle={Proceedings of the IEEE/CVF conference on computer vision and pattern recognition},
  pages={14765--14775},
  year={2022}
}

@inproceedings{ivanovic2023expanding,
  title={Expanding the deployment envelope of behavior prediction via adaptive meta-learning},
  author={Ivanovic, Boris and Harrison, James and Pavone, Marco},
  booktitle={2023 IEEE International Conference on Robotics and Automation (ICRA)},
  pages={7786--7793},
  year={2023},
  organization={IEEE}
}

@inproceedings{zhou2022hivt,
  title={Hivt: Hierarchical vector transformer for multi-agent motion prediction},
  author={Zhou, Zikang and Ye, Luyao and Wang, Jianping and Wu, Kui and Lu, Kejie},
  booktitle={Proceedings of the IEEE/CVF Conference on Computer Vision and Pattern Recognition},
  pages={8823--8833},
  year={2022}
}

@inproceedings{zhou2023query,
  title={Query-centric trajectory prediction},
  author={Zhou, Zikang and Wang, Jianping and Li, Yung-Hui and Huang, Yu-Kai},
  booktitle={Proceedings of the IEEE/CVF Conference on Computer Vision and Pattern Recognition},
  pages={17863--17873},
  year={2023}
}

@inproceedings{chang2019argoverse,
  title={Argoverse: 3d tracking and forecasting with rich maps},
  author={Chang, Ming-Fang and Lambert, John and Sangkloy, Patsorn and Singh, Jagjeet and Bak, Slawomir and Hartnett, Andrew and Wang, De and Carr, Peter and Lucey, Simon and Ramanan, Deva and others},
  booktitle={Proceedings of the IEEE/CVF conference on computer vision and pattern recognition},
  pages={8748--8757},
  year={2019}
}

@article{nichol2018first,
  title={On first-order meta-learning algorithms},
  author={Nichol, A},
  journal={arXiv preprint arXiv:1803.02999},
  year={2018}
}

@article{vasudevan2024planning,
  title={Planning with Adaptive World Models for Autonomous Driving},
  author={Vasudevan, Arun Balajee and Peri, Neehar and Schneider, Jeff and Ramanan, Deva},
  journal={arXiv preprint arXiv:2406.10714},
  year={2024}
}

@article{wang2020tent,
  title={Tent: Ful ly test-time adaptation by entropy minimization},
  author={Wang, Dequan and Shelhamer, Evan and Liu, Shaoteng and Olshausen, Bruno and Darrell, Trevor},
  journal={arXiv preprint arXiv:2006.10726},
  year={2020}
}

@inproceedings{houenou2013vehicle,
  title={Vehicle trajectory prediction based on motion model and maneuver recognition},
  author={Houenou, Adam and Bonnifait, Philippe and Cherfaoui, V{\'e}ronique and Yao, Wen},
  booktitle={2013 IEEE/RSJ international conference on intelligent robots and systems},
  pages={4363--4369},
  year={2013},
  organization={IEEE}
}

@inproceedings{hu2023planning,
  title={Planning-oriented autonomous driving},
  author={Hu, Yihan and Yang, Jiazhi and Chen, Li and Li, Keyu and Sima, Chonghao and Zhu, Xizhou and Chai, Siqi and Du, Senyao and Lin, Tianwei and Wang, Wenhai and others},
  booktitle={Proceedings of the IEEE/CVF Conference on Computer Vision and Pattern Recognition},
  pages={17853--17862},
  year={2023}
}

@article{li2017meta,
  title={Meta-sgd: Learning to learn quickly for few-shot learning},
  author={Li, Zhenguo and Zhou, Fengwei and Chen, Fei and Li, Hang},
  journal={arXiv preprint arXiv:1707.09835},
  year={2017}
}

@article{javed2019meta,
  title={Meta-learning representations for continual learning},
  author={Javed, Khurram and White, Martha},
  journal={Advances in neural information processing systems},
  volume={32},
  year={2019}
}

@inproceedings{sun2020scalability,
  title={Scalability in perception for autonomous driving: Waymo open dataset},
  author={Sun, Pei and Kretzschmar, Henrik and Dotiwalla, Xerxes and Chouard, Aurelien and Patnaik, Vijaysai and Tsui, Paul and Guo, James and Zhou, Yin and Chai, Yuning and Caine, Benjamin and others},
  booktitle={Proceedings of the IEEE/CVF conference on computer vision and pattern recognition},
  pages={2446--2454},
  year={2020}
}

@inproceedings{houston2021one,
  title={One thousand and one hours: Self-driving motion prediction dataset},
  author={Houston, John and Zuidhof, Guido and Bergamini, Luca and Ye, Yawei and Chen, Long and Jain, Ashesh and Omari, Sammy and Iglovikov, Vladimir and Ondruska, Peter},
  booktitle={Conference on Robot Learning},
  pages={409--418},
  year={2021},
  organization={PMLR}
}

@article{zhan2019interaction,
  title={Interaction dataset: An international, adversarial and cooperative motion dataset in interactive driving scenarios with semantic maps},
  author={Zhan, Wei and Sun, Liting and Wang, Di and Shi, Haojie and Clausse, Aubrey and Naumann, Maximilian and Kummerle, Julius and Konigshof, Hendrik and Stiller, Christoph and de La Fortelle, Arnaud and others},
  journal={arXiv preprint arXiv:1910.03088},
  year={2019}
}

@inproceedings{caesar2020nuscenes,
  title={nuscenes: A multimodal dataset for autonomous driving},
  author={Caesar, Holger and Bankiti, Varun and Lang, Alex H and Vora, Sourabh and Liong, Venice Erin and Xu, Qiang and Krishnan, Anush and Pan, Yu and Baldan, Giancarlo and Beijbom, Oscar},
  booktitle={Proceedings of the IEEE/CVF conference on computer vision and pattern recognition},
  pages={11621--11631},
  year={2020}
}

@inproceedings{park2024improving,
  title={Improving transferability for cross-domain trajectory prediction via neural stochastic differential equation},
  author={Park, Daehee and Jeong, Jaewoo and Yoon, Kuk-Jin},
  booktitle={Proceedings of the AAAI Conference on Artificial Intelligence},
  volume={38},
  number={9},
  pages={10145--10154},
  year={2024}
}

@inproceedings{ivanovic2022propagating,
  title={Propagating state uncertainty through trajectory forecasting},
  author={Ivanovic, Boris and Lin, Yifeng and Shrivastava, Shubham and Chakravarty, Punarjay and Pavone, Marco},
  booktitle={2022 International Conference on Robotics and Automation (ICRA)},
  pages={2351--2358},
  year={2022},
  organization={IEEE}
}

@inproceedings{salzmann2020trajectron++,
  title={Trajectron++: Dynamically-feasible trajectory forecasting with heterogeneous data},
  author={Salzmann, Tim and Ivanovic, Boris and Chakravarty, Punarjay and Pavone, Marco},
  booktitle={Computer Vision--ECCV 2020: 16th European Conference, Glasgow, UK, August 23--28, 2020, Proceedings, Part XVIII 16},
  pages={683--700},
  year={2020},
  organization={Springer}
}

@inproceedings{zhou2024smartrefine,
  title={Smartrefine: A scenario-adaptive refinement framework for efficient motion prediction},
  author={Zhou, Yang and Shao, Hao and Wang, Letian and Waslander, Steven L and Li, Hongsheng and Liu, Yu},
  booktitle={Proceedings of the IEEE/CVF Conference on Computer Vision and Pattern Recognition},
  pages={15281--15290},
  year={2024}
}

@article{park2023leveraging,
  title={Leveraging future relationship reasoning for vehicle trajectory prediction},
  author={Park, Daehee and Ryu, Hobin and Yang, Yunseo and Cho, Jegyeong and Kim, Jiwon and Yoon, Kuk-Jin},
  journal={arXiv preprint arXiv:2305.14715},
  year={2023}
}

@inproceedings{finn2017model,
  title={Model-agnostic meta-learning for fast adaptation of deep networks},
  author={Finn, Chelsea and Abbeel, Pieter and Levine, Sergey},
  booktitle={International conference on machine learning},
  pages={1126--1135},
  year={2017},
  organization={PMLR}
}

@inproceedings{ye2023improving,
  title={Improving the generalizability of trajectory prediction models with frenet-based domain normalization},
  author={Ye, Luyao and Zhou, Zikang and Wang, Jianping},
  booktitle={2023 IEEE International Conference on Robotics and Automation (ICRA)},
  pages={11562--11568},
  year={2023},
  organization={IEEE}
}

@inproceedings{park2024t4p,
  title={T4P: Test-Time Training of Trajectory Prediction via Masked Autoencoder and Actor-specific Token Memory},
  author={Park, Daehee and Jeong, Jaeseok and Yoon, Sung-Hoon and Jeong, Jaewoo and Yoon, Kuk-Jin},
  booktitle={Proceedings of the IEEE/CVF Conference on Computer Vision and Pattern Recognition},
  pages={15065--15076},
  year={2024}
}
}

\end{document}